\documentclass{article}

\usepackage[square,sort,comma,numbers]{natbib}
\usepackage[nonatbib,preprint]{neurips_2023}

\usepackage[utf8]{inputenc} 
\usepackage[T1]{fontenc}    
\usepackage{hyperref}       
\usepackage{url}            
\usepackage{booktabs}       
\usepackage{amsfonts}       
\usepackage{nicefrac}       
\usepackage{microtype}      
\usepackage{graphics,graphicx} 
\usepackage[shortlabels]{enumitem}
\usepackage{graphicx,wrapfig}
\usepackage{array, makecell}%
\setcellgapes{4pt}
\usepackage{makecell}
\usepackage{graphicx}
\usepackage{subcaption}
\usepackage[dvipsnames]{xcolor}
\usepackage{multirow}

\usepackage{graphics,graphicx}

\usepackage[shortlabels]{enumitem}
\usepackage{algorithm}
\usepackage{algpseudocode}
\usepackage{array, makecell}%
\setcellgapes{4pt}

\usepackage{amsmath}
\algrenewcommand\algorithmicrequire{\textbf{Input:}}
\algrenewcommand\algorithmicensure{\textbf{Output:}}

\setcitestyle{square}

\title{Improving Molecular Properties Prediction Through Latent Space Fusion}

\author{%
  Eduardo Soares\\
  IBM Research  - Brazil \\
  \texttt{eduardo.soares@ibm.com} \\
  \And
Akihiro Kishimoto \\
  IBM Research - Japan \\
  \texttt{akihiro.kishimoto@ibm.com} \\
  \And
  Emilio Vital Brazil \\
  IBM Research - Brazil \\
  \texttt{evital@br.ibm.com} \\
  \And
  Seiji Takeda \\
  IBM Research - Japan \\
  \texttt{seijitkd@jp.ibm.com} \\
\And
  Hiroshi Kajino \\
  IBM Research - Japan \\
  \texttt{kajino@jp.ibm.com} \\
  \And
  Renato Cerqueira\\
  IBM Research - Brazil \\
  \texttt{rcerq@br.ibm.com} \\
}

\begin{document}

\maketitle

\begin{abstract}

Pre-trained Language Models have emerged as promising tools for predicting molecular properties, yet their development is in its early stages, necessitating further research to enhance their efficacy and address challenges such as generalization and sample efficiency. In this paper, we present a multi-view approach that combines latent spaces derived from state-of-the-art chemical models. Our approach relies on two pivotal elements: the embeddings derived from MHG-GNN, which represent molecular structures as graphs, and MoLFormer embeddings rooted in chemical language. The attention mechanism of MoLFormer is able to identify relations between two atoms even when their distance is far apart, while the GNN of MHG-GNN can more precisely capture relations among multiple atoms closely located. In this work, we demonstrate the superior performance of our proposed multi-view approach compared to existing state-of-the-art methods, including MoLFormer-XL, which was trained on 1.1 billion molecules, particularly in intricate tasks such as predicting clinical trial drug toxicity and inhibiting HIV replication. We assessed our approach using six benchmark datasets from MoleculeNet, where it outperformed competitors in five of them. Our study highlights the potential of latent space fusion and feature integration for advancing molecular property prediction. In this work, we use small versions of MHG-GNN and MoLFormer, which opens up an opportunity for further improvement when our approach uses a larger-scale dataset.

\end{abstract}

\section{Introduction}

Chemical-based machine learning has gained widespread adoption as an efficient and accurate approach for predicting molecular properties, owing to its capacity to effectively represent crucial structural aspects of molecules \cite{fang2022geometry, wieder2020compact, shen2019molecular}. Recent advancements in foundational models have shown promising results by leveraging chemical language representations through a two-step process of pre-training on extensive unlabeled corpora and subsequent fine-tuning on specific downstream tasks of interest \cite{takeda2023foundation, soares2023beyond, horawalavithana2022foundation}.

Despite the emergence of pre-trained Language Models as viable options for molecular property prediction \cite{white2023future, pan2023large, white2022large}, they are still in their nascent stages of development. There is a pressing need for further research to enhance their performance and address issues like generalization and sample efficiency \cite{wang2019smiles,moret2022perplexity}.

Furthermore, recent discussions have emphasized the pivotal role of enhancing data quality and representations in elevating the overall quality of generated models and diminishing the reliance on excessively large models \cite{gunasekar2023textbooks, eldan2023tinystories}. The incorporation of high-quality data and representations has the potential to advance the state-of-the-art in pre-trained Language Models while concurrently decreasing the dataset volume and training computational resources \cite{gunasekar2023textbooks}. Of equal significance, the adoption of smaller models that necessitate less extensive training can substantially mitigate the environmental impact associated with large pre-trained Language Models \cite{gunasekar2023textbooks}.

In this study, we introduce a multi-view approach that leverages on the fusion of latent spaces from different natures generated by two state-of-the-art chemical-based models, namely MoLFormer-base \cite{ross2022large} which is based on Transformers, and MHG-GNN a graph-based approach. Our approach is geared towards enhancing the prediction of molecular properties. Our findings  demonstrate that our proposed method surpasses existing state-of-the-art algorithms, including the chemical language-based MoLFormer-XL, when it comes to tackling intricate tasks like predicting the toxicity of drugs in clinical trials and gauging the potential of small molecules to inhibit HIV replication. These challenging tasks are part of the MoleculeNet benchmark dataset \cite{wu2018moleculenet}. Furthermore, our approach exhibits superior performance in five out of six datasets studied during our experiments.

It is also important to highlight that the proposed approach refers to a fusion of latent spaces of models that were trained on 1.7 million molecules (combined), and consistently performed better than MoLFormer-XL which was trained on 1.1 billion molecules. This prompts the discussion of the necessity of very large models which are costly, resource-hungry, and laborious. In fact, our approach opens up promising avenues for future research in molecular property prediction. By leveraging the fusion of latent spaces and feature sets, we have demonstrated a significant enhancement in performance that holds potential for advancing the field.

\section{Methodology}

\begin{figure}[H]
    \centering
    \includegraphics[width=14cm]{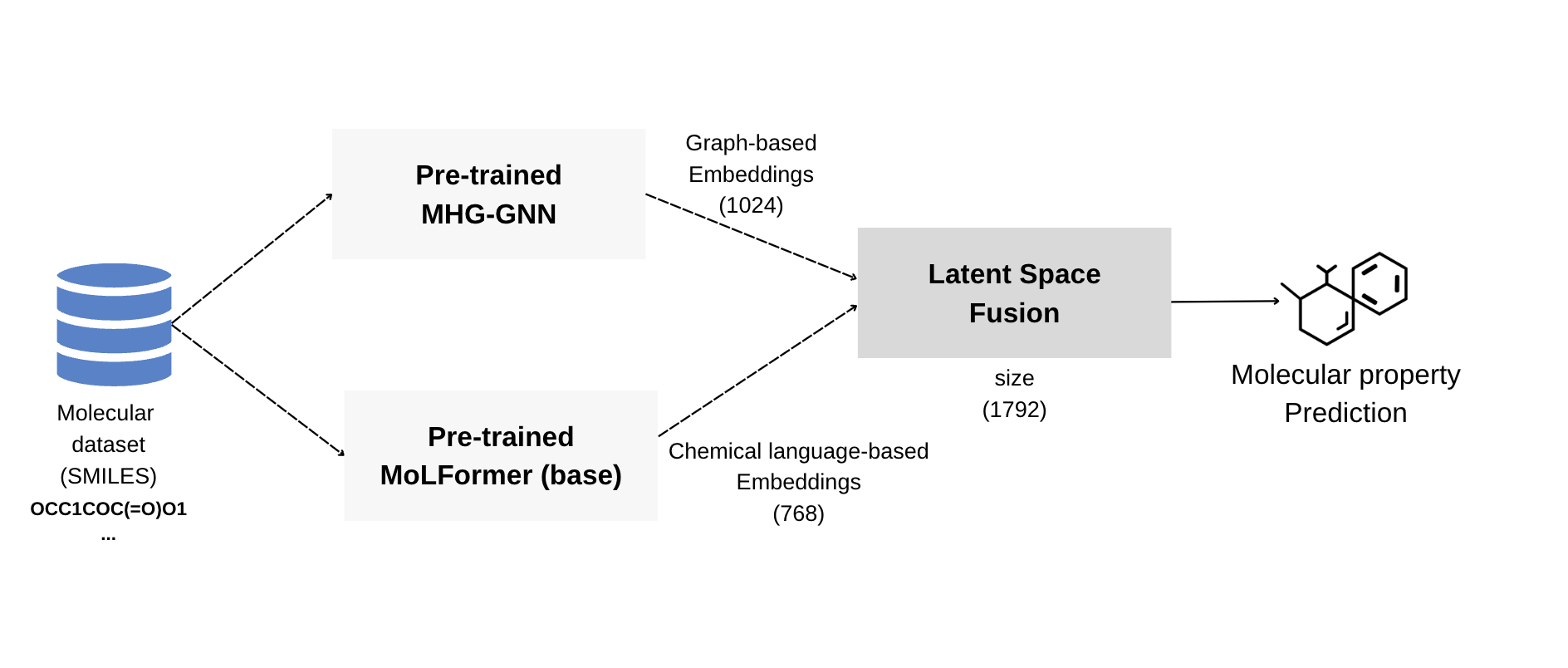}
    \caption{
General architecture of the proposed multi-view  approach.
    }
    \label{fig:proposal}
\end{figure}

In this section, we explain the methodological framework delineated within this study. As depicted in Figure \ref{fig:proposal}, we present an intricately devised schema for latent space fusion. Our approach relies on two pivotal elements: the embeddings derived from MHG-GNN, which represent molecular structures as graphs, and MoLFormer embeddings rooted in chemical language. This fusion of latent spaces is used for downstream tasks aiming at the prediction of molecular properties.  

Our approach combines two orthogonal embeddings. A GNN architecture of MHG-GNN can more accurately capture molecular substructures than MoLFormer. On the other hand, a self-attention mechanism of MoLFormer has advantage of accounting for a relation between one atom to the other atoms even if their distances are larger than the radius of GNN. 

We extract the embeddings for each SMILES contained in the dataset that we are exploring based on the pre-trained networks. For the MHG-GNN the embeddings space has the size of 1024, for MoLFormer-base the embeddings size is 768. Then, the resulting fused latent space has the size of 1792. Details of the employed models are described in the next subsections. XGBoost \cite{chen2015xgboost} with \textit{optuna} \cite{akiba2019optuna} optimizer was employed as predictor.

\subsection{MHG-GNN}

MHG-GNN \cite{kishimoto2023mhggnn} is an autoencoder that combines GNN with Molecular Hypergraph Grammar (MHG) introduced for MHG-VAE \cite{Kajino:ICML2019}.
Unlike existing autoencoders that receive their input and output in the same format, MHG-GNN receives them in a different format.
MHG-GNN receives a molecular structure represented as a graph. The encoder constructed as Graph Isomorphism Network \cite{Xu:ICLR2019} that additionally considers edges encodes that graph to its corresponding latent vector \cite{Hu:ICLR2020}.
The decoder constructed as GRU and with several neural network models decodes that latent vector to the original molecular structure represented as a sequence of production rules on molecular hypergraphs.
The production rules are generated from the dataset for pre-training. 

MHG-GNN can inherit advantage of MHG-VAE that can always generate structurally valid molecular structures when decoding latent vectors.
Additionally, MHG-GNN can always embed graph structures to their latent vectors, whereas the encoder of MHG-VAE cannot always; it cannot accept a molecule that cannot be represented by a set of production rules generated from the dataset for pretraining. 
Finally, thanks to GNN, MHG-GNN has more direct understanding to the structural information than language-based models, which may capture different characteristics than MoLFormer. 

We used the model trained in the same steps described in \cite{kishimoto2023mhggnn} and with a radius of 3 (i.e., the iteration size for message passing step in GNN). With these configurations, MHG-GNN generates 1024 embeddings. 
MHG-GNN was pre-trained on 1,381,747 molecules extracted from the PubChem database in its training part, this process generates 16,362 rules that represent these molecules.

\subsection{MoLFormer}

MoLFormer \cite{ross2022large}, is a large-scale masked chemical language model that processes inputs through a series of blocks that alternate between self-attention and feed-forward connections. MoLFormer was trained in a self-supervision manner with 1.1 billion molecules from PubChem and ZINC datasets and uses tokenization process, as detailed in \cite{schwaller2019molecular}. The MoLFormer vocabulary includes 2362 unique chemical tokens. These tokens are used to fine-tune or retrain the MolFormer model. To reduce computation time, the sequence length has been limited to a range of 202 tokens as 99.4\% percent of all 1.1 billion molecules contain less than 202 tokens.

\textsc{MoLFormer} is equipped with a self-attention mechanism that allows the network to construct complex representations that incorporate context from across the sequence of SMILES. By transforming the sequence features into queries ($q$), keys ($k$), and value ($v$) representations, attention mechanisms can weigh the importance of different elements within the sequence. 
This enables the model to learn highly informative representations of the input data, making it a powerful tool for predicting molecular properties.

In this work, we used the base version of the MoLFormer that was trained on a small portion of molecules compared to the MoLFormer-XL version. The MoLFormer-base version it is publicly available at \url{https://github.com/IBM/molformer}.

\section{Results}

To assess the efficacy of our proposed methodology, we conducted comprehensive experiments using six distinct benchmark datasets. These datasets encompass a wide spectrum of tasks, ranging from predicting the physical and biophysical properties to the physiological attributes of small molecule chemicals. The datasets were sourced from MoleculeNet \cite{wu2018moleculenet}, and their characteristics are summarized in Table \ref{TableDataset}. To gauge the performance of our models, we employed AUC-ROC as the evaluation metric, employing scaffold splits for the assessments. For a fair comparison, the train/validation/test split employed here was the same as \cite{ross2022large}.

\begin{table}[htbp]
\scriptsize
	\begin{center}
		\caption{Benchmark datasets from MoleculeNet}\label{TableDataset}
		\begin{tabular}{cc|c}
        
			\hline
              Dataset & Description &  number samples \\
			\hline
		
			BBBP & Blood brain barrier penetration dataset  & 2039 \\
                Tox21 & Toxicity measurements on 12 different targets  & 7831 \\
                Clintox & Clinical trial toxicity of drugs  & 1478 \\
                HIV &  Ability of small molecules to inhibit HIV replication  & 41127 \\
                BACE &  Binding results for a set of inhibitors for $\beta$ – secretase 1  & 1513 \\
                SIDER &  Drug side effect on different organ classes  & 1427 \\
			
			\hline
		\end{tabular}
	\end{center}
\end{table}

Table \ref{tab:methods_performance} offers a comprehensive overview of the comparative performance between our proposed multi-view approach and state-of-the-art algorithms on various benchmark datasets. A keen analysis of the table reveals that the multi-view approach, which leverages the fusion of embeddings, outperforms its counterparts in 5 out of 6 datasets, underscoring its potential to excel in diverse domains. Particularly noteworthy is the consistent superiority of our fusion-based approach over the MHG-GNN and MoLFormer-Base methods in all conducted experiments. This pattern of success across multiple datasets strongly suggests that the fusion of embeddings from different natures plays a pivotal role in enhancing the algorithm's performance.

\begin{table}[h]
\centering
\caption{Methods and Performance}
\begin{tabular}{lcccccc}
\hline
\multirow{2}{*}{Method} & \multicolumn{6}{c}{Dataset} \\
\cline{2-7}
 & BBBP & ClinTox & HIV & BACE & SIDER & Tox21 \\
\hline
RF\cite{ross2022large} & 71.4 & 71.3 & 78.1 & 86.7 & 68.4 & 76.9 \\
SVM\cite{ross2022large} & 72.9 & 66.9 & 79.2 & 86.2 & 68.2 & 81.8 \\
MGCN\cite{lu2019molecular} & 85.0 & 63.4 & 73.8 & 73.4 & 55.2 & 70.7 \\
D-MPNN\cite{yang2019analyzing} & 71.2 & 90.5 & 75.0 & 85.3 & 63.2 & 68.9 \\
DimeNet\cite{gasteiger2020directional} & - & 76.0 & - & - & 61.5 & 78.0 \\
Hu, et al.\cite{hu2019strategies} & 70.8 & 78.9 & 80.2 & 85.9 & 65.2 & 78.7 \\
N-Gram\cite{liu2019n} & 91.2 & 85.5 & 83.0 & 87.6 & 63.2 & 76.9 \\
MolCLR\cite{wang2022molecular} & 73.6 & 93.2 & 80.6 & 89.0 & 68.0 & 79.8 \\
GraphMVP\cite{liu2021pre} & 72.4 & 77.5 & 77.0 & 81.2 & 63.9 & 74.4 \\
GeomGCL\cite{liu2021pre} & - & 91.9 & - & - & 64.8 & \textbf{85.0} \\
GEM\cite{fang2022geometry}& 72.4 & 90.1 & 80.6 & 85.6 & 67.2 & 78.1 \\
ChemBerta\cite{chithrananda2020chemberta} & 64.3 & 90.6 & 62.2 & - & - & - \\
MoLFormer-XL\cite{ross2022large} & 93.7 & 94.8 & 82.2 & 88.21 & 69.0 & 84.7 \\
MoLFormer-Base\cite{ross2022large}  & 90.9 & 77.7 & 82.8 & 64.8 & 61.3 & 43.2 \\
MHG-GNN  & 93.48 & 89.97 & 83.40 & 87.29 & 67.62 & 77.50 \\
\textbf{Latent Space fusion}\\ (MHG-GNN and MoLFormer-Base) & \textbf{94.19} & \textbf{98.75} & \textbf{86.08} & \textbf{90.37} & \textbf{69.88} & 80.46 \\
\hline
\end{tabular}
\label{tab:methods_performance}
\end{table}

Our proposed fusion-based approach harnesses the power of 768 embeddings from transformers-based MoLFormer-Base and 1024 embeddings from graph-based MHG-GNN, capitalizing on their complementary strengths to excel in a variety of challenging tasks. 

It is important to highlight that we use the fusion of latent spaces of two small models when compared to the state-of-the-art, MoLFormer-base and MHG-GNN was pre-trained in a small portion of selected molecules from PubChem. The fusion of these smalls performed better than MoLFormer-XL which was trained in 1.1 billion molecules in 5 out of 6 benchmarks datasets. This not only highlights our method's effectiveness but also  paves the way for additional enhancements when our approach leverages a larger-scale dataset.

In conclusion, the findings showcased in Table \ref{tab:methods_performance} underscore the potential of our proposed multi-view approach, emphasizing its ability to harness diverse features for improved performance in various complex tasks. Future experiments should focus on different fusion strategies, as well as better quality features/embeddings. 

\section{Conclusion}

This paper introduces a multi-view approach that synergizes the latent spaces of two state-of-the-art algorithms for molecules properties prediction, MoLFormer-base and MHG-GNN, to deliver superior performance across a range of demanding tasks sourced from the MoleculeNet dataset. Our proposed method has consistently outperformed the state-of-the-art competitors on 5 out of 6 benchmark datasets, showcasing its robustness and versatility. 

The proposed multi-view approach used features from models trained on small datasets and performed consistently better than MoLFormer-XL, state-of-the-art, which was trained on 1.1 billion molecules. This opens up an opportunity for further improvement when our approach uses a larger-scale dataset.

Future research direction will focus on the exploration of diverse fusion techniques and the integration of high-quality features and embeddings for refining our approach. These findings offer a promising direction for enhancing the accuracy and effectiveness of molecular property prediction.

\footnotesize
\bibliographystyle{IEEEtran}
\bibliography{sample}

\newpage
\section*{Supplementary Materials}

\subsection*{ROC-AUC curves for single task datasets}

In this section, we present ROC-AUC curves for three single-task datasets: BBBP, HIV, and BACE. These curves serve as a visual representation of our proposed model's ability to discriminate between two classes at varying threshold levels. A higher AUC value corresponds to superior classification performance.

\begin{figure}[H]
    \centering
    \includegraphics[width=12cm]{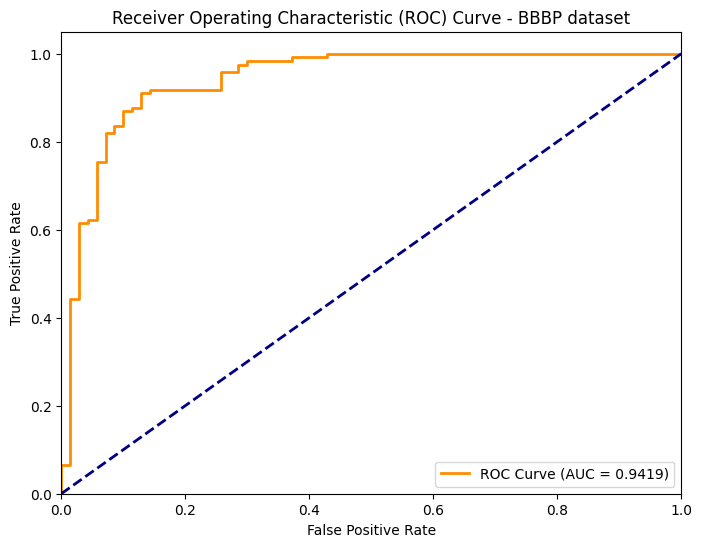}
    \caption{
ROC-AUC curve for the BBBP dataset.
    }
    \label{fig:output_BBBP}
\end{figure}

Fig \ref{fig:output_BBBP} illustrates the ROC-AUC curve for the BBBP dataset. The proposed multi-view approach obtained an AUC score of $94.19$, surpassing the state-of-the-art for this task which was the MoLFormer-XL with an AUC score of $93.7$.

\begin{figure}[H]
    \centering
    \includegraphics[width=12cm]{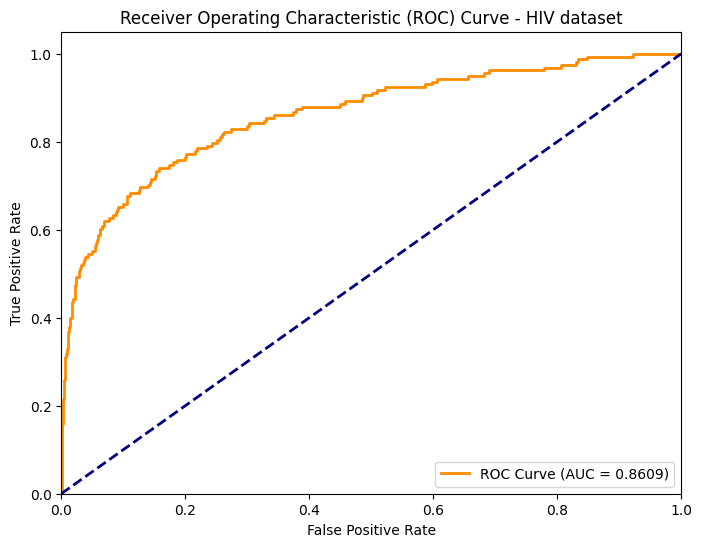}
    \caption{
ROC-AUC curve for the HIV dataset.
    }
    \label{fig:output_HIV_ROC}
\end{figure}

Fig \ref{fig:output_HIV_ROC} illustrates the ROC-AUC curve for the HIV dataset. The HIV dataset  involve predicting whether a compound has anti-HIV activity. This can be crucial in the development of antiretroviral drugs. The proposed multi-view approach obtained an AUC score of $86.09$, which is better than the other state-of-the-art algorithms.

\begin{figure}[H]
    \centering
    \includegraphics[width=12cm]{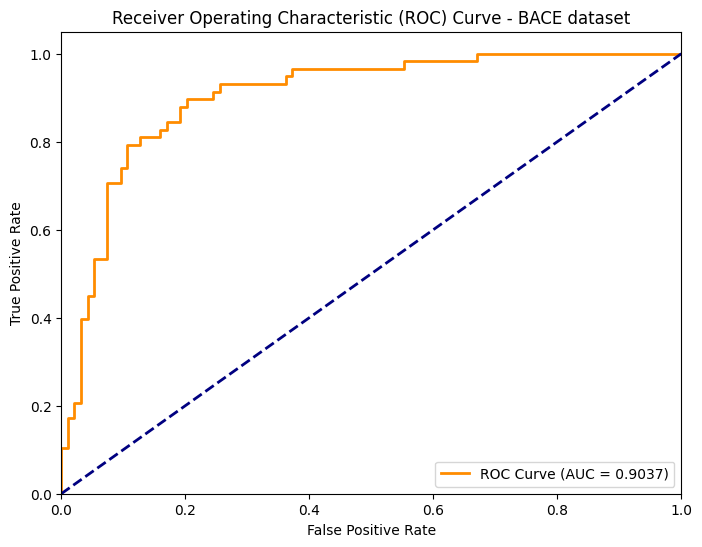}
    \caption{
ROC-AUC curve for the BACE dataset.
    }
    \label{fig:output_BACE}
\end{figure}

The BACE dataset refers to an enzyme involved in the formation of beta-amyloid plaques, which are associated with Alzheimer's disease \cite{wu2018moleculenet}. Therefore, the BACE dataset involves the prediction of whether a compound inhibits or activates this enzyme, which can be relevant for Alzheimer's drug development \cite{wu2018moleculenet}. Fig. \ref{fig:output_BACE} illustrates the ROC-AUC curve for the proposed approach to this task. The proposed approach obtained an AUC score of $90.37$ which is better than the current state-of-the-art approaches.

\end{document}